# An Algorithm for Aligning Sentences in Bilingual Corpora
# Using Lexical Information

**Akshar Bharati,**
**V.Sriram, A.Vamshi Krishna,**
**Rajeev Sangal, Sushma Bendre**
**International Institute of Information Technology,**
**Hyderabad.**
**{sriram,vamshi_a}@gdit.iiit.net,{sangal,bendre}@iiit.net**

# An algorithm for Aligning Sentences in Bilingual Corpora Using Lexical information

## Abstract

In this paper we describe an algorithm for aligning sentences with their translations in a bilingual corpus using lexical information of the languages. Existing efficient algorithms ignore word identities and consider only the sentence lengths (Brown, 1991; Gale and Church, 1993). For a sentence in the source language text, the proposed algorithm picks the most likely translation from the target language text using lexical information and certain heuristics. It does not do statistical analysis using sentence lengths. The algorithm is language independent. It also aids in detecting addition and deletion of text in translations. The algorithm gives comparable results with the existing algorithms in most of the cases while it does better in cases where statistical algorithms do not give good results.

## 1. Introduction

Aligned bilingual corpora have proved useful in many ways including machine translation, sense disambiguation and bilingual lexicography. The task of alignment has proved to be difficult in many ways. For some languages, it is difficult to use the statistical analysis of sentence lengths to do the alignment. Further, there are substantial additions and deletions that can occur on either side, particularly when the languages are far apart . A lot of sentences align many to many and this makes the task more difficult.

There are a few existing algorithms which do good alignment. One such algorithm is the Gale and Church Algorithm (1993). The Gale and Church Algorithm is basically dependent on the length of the sentence in terms of characters and the Brown's algorithm (1991) is dependent on the length of the sentence in terms of words. Dynamic programming is then used to search for the best alignment in both the algorithms. These algorithms cannot be used effectively to align a very large corpus taken as a single unit. Therefore it depends heavily on the paragraph delimiters, which are called 'hard delimiters'. These paragraph delimiters also help the Gale and Church algorithm to correct itself if it is going wrong. These paragraph markers sometimes may not be present in the corpus. Also the source text paragraphs and the target text paragraph may not align with each other. For example the parallel-corpus that we used had no paragraph delimiters. To use the Gale and Church algorithm on the parallel-corpora with no paragraph delimiters, the delimiters have to be introduced manually.

While both the Gale and Church algorithm and Brown's algorithm have achieved remarkably good outputs for language pairs like English-French and English-German with error rates of 4% on an average, there is still a great scope for improvement.  The algorithm is not robust with respect to non-literal translations and deletions. Also with an algorithm, which relies only on length of the sentence, it is quite difficult to automatically recover from large deletions.

The Gale and Church did not work well on the parallel-corpus that we have, and the need to use the lexical information to do the alignment was felt. Also alignment algorithms that use lexical information offer a potential for high accuracy on any corpus. Chen (1993) did considerable amount of work on English-French corpus using lexical information.

The algorithm that is proposed in this paper does the sentence alignment at a high level of accuracy using lexical information of the corpus and available lexical resources of the source and target languages. One such resource used is the bilingual lexicon. The proposed algorithm uses a medium-coverage dictionary (25,000 words lexicon) to do the alignment. The other resources that are used include the chunkers for both the languages. The algorithm first breaks the sentences of both the languages into small units called chunks. To find an alignment for a sentence in the source language, it is matched with a set of possible sentences in the target language and the scores are assigned for each comparision. The score of match of two sentences is calculated by finding out the number of chunks that match between the two sentences. The algorithm then carries out the alignment of sentences using these scores. The precision of the alignment is 94.3%.

# 2. Background

## 2.1 Parallel Bilingual Corpus:

In this section, we describe the data that we used to test the algorithm. The data comes from a weekly news magazine "India-Today". The magazine is released in two languages. The source language is English and the target language into which it is later translated is Hindi. English is a fixed-word order language while Hindi is a free-word order language. Free-word order languages are those where the order of words can be changed without losing the meaning. Hence, the sentence lengths of both the languages are not proportional which makes it difficult to use statistical analysis to do the alignment. Also, there are substantial additions and deletion of texts in either of the issues of the magazine.

## 2.2 Framework:

A text in a source language and the corresponding text in a target language are given to the alignment system. First, all the source sentences and the target sentences are chunked into smaller units based on the language specific chunkers. Our aim is to identify an appropriate translation for a particular sentence in the source language text among the sentences in the target language text. To do this, the source sentence is first compared with a set of probable sentences that could be the translation of the source sentence. A score of comparison is assigned for every such matching.

The score of comparison between a source sentence and a target sentence is determined by comparing the chunks of both the sentences using a English-Hindi lexicon. To assign the scores, we identify the number of chunks of the source that are actual translations in the target language. Based on number of matching chunks out of available chunks, various scoring functions can be used.

The figure below gives an overview of the Sentence alignment System

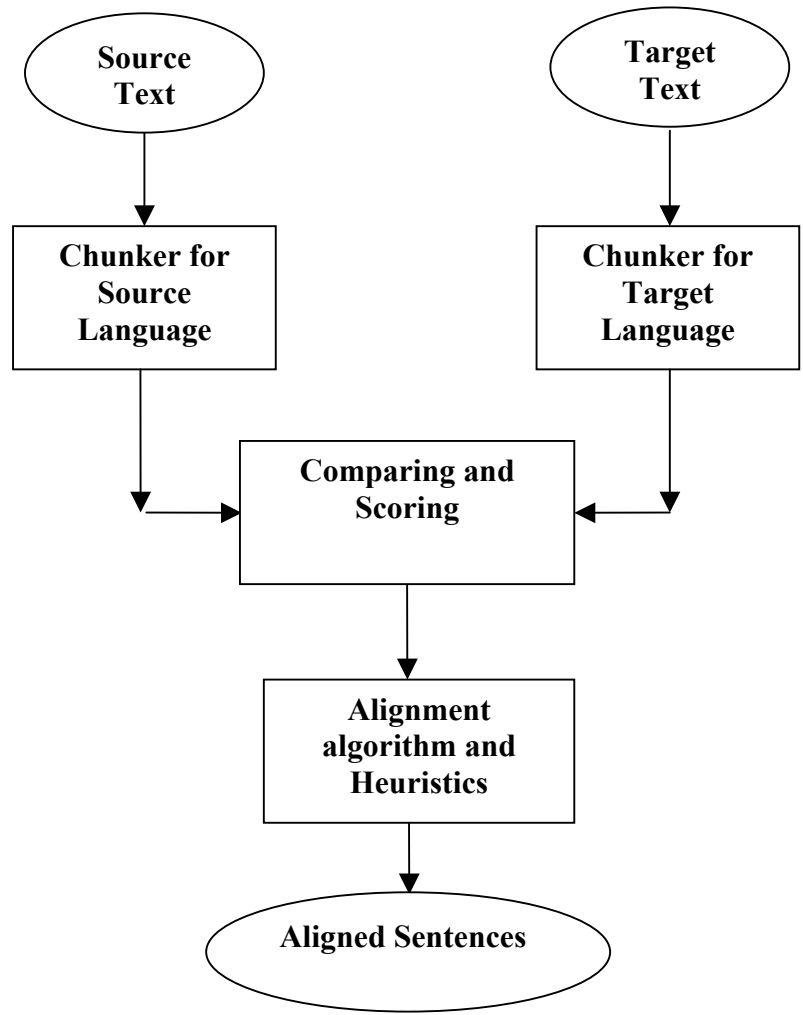

**Figure 1. Summarizing the framework of the algorithm**

As depicted in the Figure 1, there are three stages in the proposed alignment algorithm namely Chunking, Scoring and Alignment.

# 3. Chunking

In this section, we explain Chunking and how it is useful to do the alignment. Chunking involves finding the groups of related words in a sentence. The chunks are used to refer to a single concept. A sentence can thus be looked at as a sequence of chunks, each chunk adding information to the sentence. Hence, the chunks can be seen as the building blocks of a sentence in conceptual terms. They are often used to do the analysis of the sentences.

Also, the lexicons may not be exhaustive for all languages. By chunking sentences, the headwords can be identified and can be used in matching where as the other words can be given a relatively lesser weight and therefore even if a dictionary provides less word matches, it is not a considerable drawback. If it were just a word-to-word match where every word has equal weight, then any word that is not supported by the lexicon of that language would mislead the alignment.

The chunks can be categorized as
- Noun chunks.
- Verb chunks.

## 3.1 Noun chunks:

A non-recursive noun phrase is a noun chunk .It typically consists of a determiner and optional adjectives followed by a noun. Prepositions preceding the noun phrases are also grouped with the noun phrase. Noun chunks are usually the same as the noun phrases.

For example,
1. 'The red party':: Here, "The" = determinant, "red" = adjective and "party " = noun.
2. 'of the cast iron pump'::  Here, "of" = preposition, "the" = determinant, "cast" and "iron" = adjectives and "pump" = noun.

## 3.2 Verb chunks:

A verb chunk consists of a group of main verb, supporting or auxiliary verbs and the adverbs. The auxiliary verbs in a sentence tell the tense, aspect and the modality of the sentence. The main verb carries the lexical information in a verb group. Verb chunks are usually same as the verb phrases.

For example:
1. 'is playing'::  Here, "playing" = main verb, "is" = supporting verb.
2. 'would have been going fast':: Here, "would have been" = auxiliary verbs, "going" = main verb and "fast" = adverb.

Sentences can be chunked as shown below.

**English:**
[The gigantic migratory fish] ((has been sought out)) [in Gujarat] [since ancient times] [for its liver oil] .

**Hindi :**
[इस विशालकाय प्रवासी  मछलि के] [जिगर के तेल के लिए] [गुजरात में] [प्राचीन काल से ही]
    gigantic  migratory  fish      liver  oil          Gujarat    ancient times

(( इसकी काफी  मांग रही है ))।
          sought

Noun chunks are enclosed in [] while the verb chunks are enclosed in (()).

## 4. Chunk Matching

Two chunks are matched by first matching their headwords and then matching the support words of a chunk. The chunk matching thus is done by looking inside a chunk, that is, the words constituting the chunks. In a Noun chunk, all the words except the prepositions and postpositions are used to do the chunk matching whereas in a Verb chunk, only the headword is used to do the chunk matching.

Two words of a chunk are matched by any of the following ways-

1. Bi-lingual dictionary lookup: - To match two words, we look at the meanings of the word of source language sentence in the dictionary and check if the target word is one among them.

2. Target-Target dictionary lookup:- Non-availability of the match of a source word and a target word in the Bi-Lingual dictionary would result in a further lookup in a target-target dictionary if available.

3. Numeric matching: - A number in a source sentence that shows a correspondence in the target sentence would result in a reliable match.

4. Phonetic matching: - If the words are proper nouns, the words are matched using a phonetic matcher.

# 5. Alignment Algorithm

The alignment is done after the scores of comparisons are assigned which are obtained by matching the chunks. Different scoring functions can be used to calculate the score of match. The following gave better results than the rest as it gives a more stable match between the two.

$$\text{Score\_of\_match}(S, T) = \frac{\text{Number of matching chunks}}{\text{Maximum (source chunks, target chunks)}}$$

S: Source sentence
T: Target sentence

The alignment algorithm takes the scores of comparison in decreasing order and considers each pair for alignment. Hence, the scores of comparisons are first sorted.
Only those pairs of sentences are considered whose scores of comparison are above a particular threshold. The value of the threshold can be any value more than 0. The threshold affects the precision and recall of the alignment. As the threshold increases, the precision increases while the recall becomes less. The process of checking whether a pair of sentences form an alignment or not is governed by few heuristics.

Note that the above algorithm gives a one-one mapping of sentences.

The heuristics that are used
1. The aligner does not match two sentences to one sentence or vice-versa.
   ie.., a cannot align with b if a already aligns with c.
   This heuristic is applied because usually the number of chunk matching is very less and would give rise to error in the absence of this heuristic.
2. The aligned pairs of sentences follow the rule of linearity,
   ie.., if a $\Leftrightarrow$ b and c $\Leftrightarrow$ d , then sign(a-c) = sign(b –d)
   This also means that the aligner does not allow any cross-linking.

Example:
Take a source text having four sentences (s1, s2, s3, s4) and a target text having two sentences (t1, t2).

Let the scores of comparison be
1. Score (s1, t1) = 5
2. Score (s2, t1) = 15
3. Score (s2, t2) = 20
4. Score (s3, s1) = 30
5. Score (s3, s2) = 7
6. Score (s4, s2) = 10

The scores are then considered in the sorted order.

1. s3 – s1      => The pair is aligned
2. s2 – t2      => This alignment gives rise to cross-mapping, hence it is not aligned. (Violation of heuristic 2)
3. s2 – s1      => The alignment violates rule 1 which says that there should only one-one mapping. Hence, this alignment is also rejected.
4. s4 – s2      => This pair is aligned.

This above example shows the working of the algorithm. The aligned pairs that are formed are (s3, s1) and (s4, s2).

## 6. Many-many alignment:

The alignment produced above can be extended to include one-many mapping. The sentences that did not get align may be a part of the one-many mapping that is possible. To be a part of one-many mapping, the sentences should be adjacent to the sentences that have already matched.

This alignment can be carried out using the following two procedures,

1. Considering number of chunks:
   The number of chunks in most of the language-pairs is proportional.
   Hence, this property can be used to attach the unaligned sentences to the already existing alignment. For example, if the number of chunks in a source sentence is 15 and the number of chunks in a target sentence are 5. This means that the target sentence is not a complete translation of the source sentence and some other sentence in the target language would complete the translation.
   Hence, the sentences adjacent to the considered are verified. The sentence that adds up more closely with the target sentence to equal the number of chunks of source sentence is also considered to be aligning with the source sentence.

2. Re-using the scores of match:
   The alignment algorithm is re-run on the sentences without disturbing the existing alignment to facilitate the many-many alignments. In this pass, we ignore the heuristic that an already aligned sentence should not be aligned with another sentence and then align the non-aligned sentences with the information of the scores of match.

# 7. Evaluation and comparison with the existing algorithms

The evaluation was done on "India-Today" corpus. About 140 texts from different issues of the magazine were taken. They included texts from diverse areas like politics, sports, business etc. The number of sentences extracted by the sentence aligner from the 140 texts are 3021, each text having an average of about 21 sentences. The sentences extracted by the system were evaluated and it was found that out of 3021 sentences extracted, 2849 sentences were aligned correctly. The precision of the system is 94.3%. The Gale and Church algorithm was run on the same set of 140 texts and was evaluated against the 2849 sentences that were correctly extracted by the proposed algorithm. It was found that the Gale and Church algorithm could identify 1767 alignments correctly out of the 2849 correct alignments. As the Gale and Church algorithm is designed to accommodate 1-2, 2-1 alignments, any partial alignment was also considered a correct alignment. The resulting precision is 62%.

The performance of the proposed algorithm can be visualized in Figure 2 by plotting the error-percentages on the horizontal axis and numbers of texts containing the corresponding error-percentages on the vertical axis. Figure 2 gives the truncated version of the graph till an error percentage of 10.

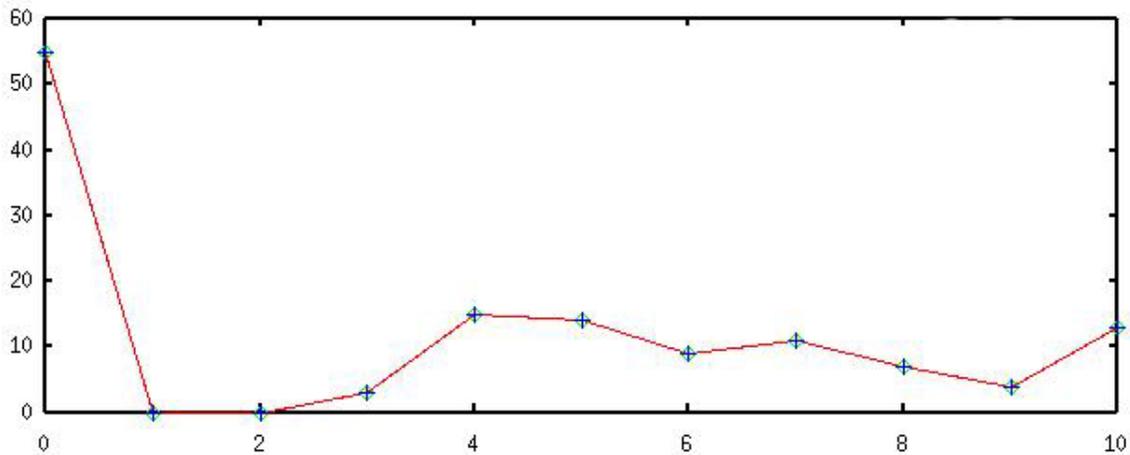

**Figure 2. Error Percentage verses Number of texts**

It can be seen from Figure 2. that out of the 140 texts used to evaluate, 55 texts gave 100% precision.

We now compare the proposed algorithm with the Gale and Church algorithm. In Figure 3., we plot text number on the horizontal axis and the precision of the text on the vertical axis. The texts were taken in the increasing order of performance of the Gale and Church algorithm. The Gale and Church algorithm gave low precisions when it is run on texts that have substantial deletions. Out of 140 texts, Gale and Church algorithm gave a precision of 100% on 22 texts as opposed to 55 texts given by the proposed algorithm.

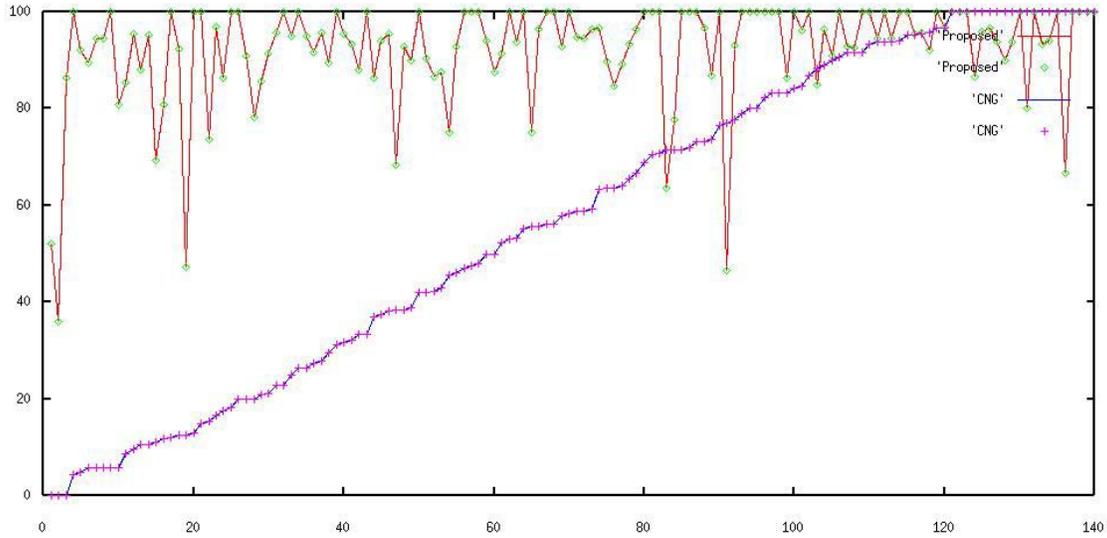
**Figure3. Comparison of Precision of Gale and Church algorithm and the proposed algorithm.**

From Figure 3, it can be seen that the Gale and Church algorithm gave 100% precision for the texts numbered from 119 to 140. Among these 22 texts, the proposed algorithm gave 100% precision for 10 texts while it gave a lower accuracy for 12 texts because of the inadequacy of the lexicon. Also, it can be seen that the Gale and Church algorithm gave a 0% precision for texts numbered from 0 to 4. On investigation, it was found that the reason for the incorrect alignment of these texts by the Gale and Church algorithm was large deletions at the beginning of these texts.

A text, which contained deletions, was aligned using the proposed algorithm and the same text was aligned using Gale and Church algorithm and the results are shown in Figure 4. For the text that was considered, the proposed alignment algorithm gave a precision of 100% while it gave a precision of only 57% when aligned using Gale and Church algorithm. The low precision of the alignment done by Gale and Church algorithm clearly depicts that it fails to align the texts that have deletions.

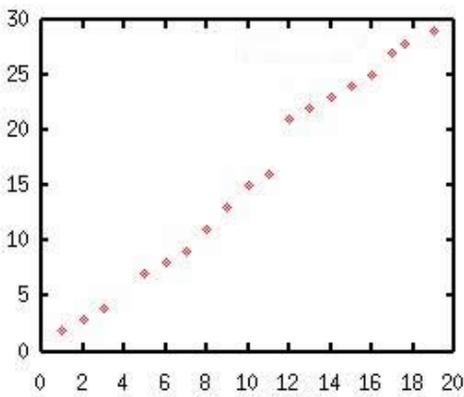 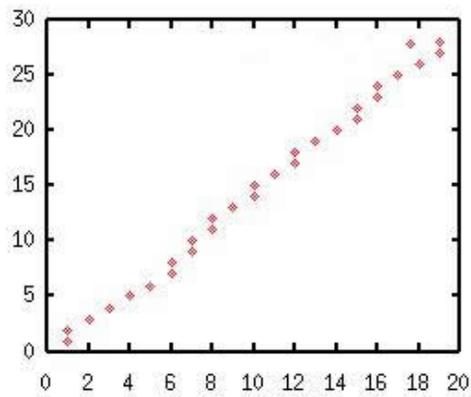

**Proposed Algorithm (Figure 4.a)**          **Gale and Church Algorithm (Figure 4.b)**

**Figure 4. Comparison of the proposed algorithm with Gale and Church algorithm for a text where the proposed algorithm does better.**

From Figure 4, we can see that the algorithm could detect a deletion of text from sentences 16 to 19, whereas the Gale and Church algorithm has failed to mark such a deletion.

The drawback of the proposed algorithm when compared to the existing algorithms is that a sentence in a source language text may sometimes not match with any of the sentences in the target language text due to the low coverage of the dictionary. This affects the recall in certain cases.

A text for which, the proposed algorithm gave a lower accuracy when compared to the Gale and Church algorithm is taken, and the results of the alignment were plotted in figure 5. The Gale and Church algorithm gave a precision of 97% while the proposed algorithm gave a precision of 87%.

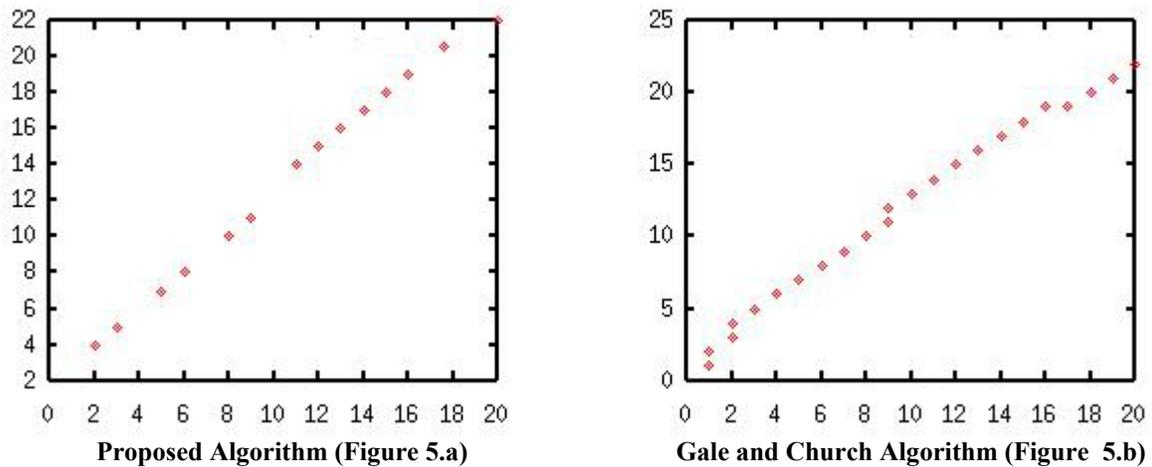

**Proposed Algorithm (Figure 5.a)**      **Gale and Church Algorithm (Figure 5.b)**

**Figure 5. Comparision of the proposed algorithm with Gale and Church algorithm for a text where the proposed algorithm does worse.**

## 8. Conclusion and Future Work

It is to be noted that the Algorithm is language independent and given the chunkers for any pair of source and target languages along with the bi-lingual lexicon can be guided to give a reasonably correct alignment of the sentences. We have done the sentence alignment as a part of our research on Example-Based Machine Translation. Evaluation of many-many alignment on the same corpus would be done in future. The next stage in an Example-Based Machine Translation system is Chunk-Alignment where we would like to increase the Chunk Matching using Heuristics and Linguistic Input. The Aligned chunks can also be used to build a Phrasal Dictionary. Also, the Aligned chunks can be used to give feedback to the lexicon used.